\documentclass{article}

\usepackage{arxiv}

\usepackage{amsmath,amsfonts,bm}

\def\eqref#1{equation~\ref{#1}}
\def\1{\bm{1}}

\DeclareMathAlphabet{\mathsfit}{\encodingdefault}{\sfdefault}{m}{sl}
\SetMathAlphabet{\mathsfit}{bold}{\encodingdefault}{\sfdefault}{bx}{n}

\usepackage{hyperref}
\usepackage{url}

\usepackage[utf8]{inputenc} % allow utf-8 input
\usepackage[T1]{fontenc}    % use 8-bit T1 fonts
\usepackage{microtype}      % microtypography
\usepackage{lipsum}		% Can be removed after putting your text content
\usepackage{natbib}
\usepackage{enumitem}
\usepackage{amsfonts}       % blackboard math symbols
\usepackage{amsmath}        % nicer math formulas 
\usepackage{nicefrac}       % compact symbols for 1/2, etc.
\usepackage{xcolor}         % colors
\usepackage{booktabs}
\usepackage{orcidlink}
\usepackage{algorithm}
\usepackage{algorithmic}
\usepackage{multirow}
\usepackage{colortbl}
\usepackage{comment}
\usepackage{cleveref}
\usepackage{subcaption}
\usepackage{xspace}
\usepackage{lscape}
\usepackage{todonotes}
\usepackage{subcaption} % put this in the preamble

\definecolor{peachfuzz}{RGB}{255, 190, 152}

\newcommand{\themodel}{Janus}

\title{Combining Euclidean and Hyperbolic Representations for Node-level Anomaly Detection}

\author{
\href{https://orcid.org/0000-0002-0961-4151}{\includegraphics[scale=0.06]{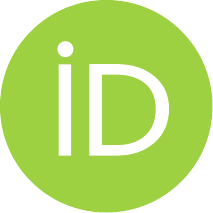}\hspace{1mm}Simone Mungari} \\
	University of Calabria\\
        ICAR-CNR\\
        Revelis s.r.l.\\
	\texttt{simone.mungari@unical.it} \\
    \And
    \href{https://orcid.org/0000-0003-3978-9291}{\includegraphics[scale=0.06]{orcid.pdf}\hspace{1mm}Ettore Ritacco} \\
	University of Udine\\
	\texttt{ettore.ritacco@uniud.it} \\
    \And 
    \href{https://orcid.org/0009-0005-9454-5910}{\includegraphics[scale=0.06]{orcid.pdf}\hspace{1mm}Pietro Sabatino} \\
	ICAR-CNR\\
	\texttt{pietro.sabatino@icar.cnr.it}
   } 
\begin{document}

\maketitle

\begin{abstract}
Node-level anomaly detection (NAD) is challenging due to diverse structural patterns and feature distributions. As such, NAD is a critical task with several applications which range from fraud detection, cybersecurity, to recommendation systems. We introduce \themodel, a framework that jointly leverages Euclidean and Hyperbolic Graph Neural Networks to capture complementary aspects of node representations. Each node is described by two views, composed by the original features and structural features derived from random walks and degrees, then embedded into Euclidean and Hyperbolic spaces. A multi Graph-Autoencoder framework, equipped with a contrastive learning objective as regularization term, aligns the embeddings across the Euclidean and Hyperbolic spaces, highlighting nodes whose views are difficult to reconcile and are thus likely anomalous. Experiments on four real-world datasets show that \themodel\ consistently outperforms shallow and deep baselines, empirically demonstrating that combining multiple geometric representations provides a robust and effective approach for identifying subtle and complex anomalies in graphs. We publicly release our source code at \url{https://anonymous.4open.science/r/JANUS-5EDF/}.
\end{abstract}

% \keywords{Data Generation, Benchmarking, Recommendation, Probabilistic Modeling}

\maketitle

\section{Introduction}\label{sec:introduction}
Anomaly detection plays a crucial role in machine learning, with impactful use cases in cybersecurity, social media analysis, financial fraud prevention, and healthcare~\citep{akoglu2015graph, west2016intelligent}. Graph anomaly detection extends this paradigm to domains where data naturally exhibits relational structure and can be modeled as a graph. Within this setting, node-level anomaly detection methods~\citep{ma2021comprehensive} focus on identifying anomalous nodes, where the notion of “anomaly” varies depending on the application. In general, anomalous nodes are those that deviate substantially from the majority in terms of their attributes or structural connectivity.

There exists a vast body of literature on graph anomaly detection, ranging from tree-based methods to deep learning frameworks. Supervised approaches~\citep{tang2023gadbench} rely on labeled anomalies; however, such labels are often scarce or unavailable, and these methods are inherently restricted to detecting only known types of anomalies. To overcome this limitation, alternative strategies employ unsupervised~\citep{bond} or semi-supervised~\citep{ioannidis2019graphsac} paradigms. Graph Neural Networks (GNNs) have emerged as the state of the art in graph representation learning, and have been extensively applied to NAD. Among the most common GNN-based strategies are graph autoencoders~\citep{dominant, fan2020anomalydae, roy2024gad} and graph contrastive learning~\citep{liu2021anomaly, pan2023prem}. Autoencoder-based methods learn to embed nodes and edges into a latent space and then reconstruct the original features and adjacency matrix. Nodes with high reconstruction error are considered anomalous, as their structure or features deviate significantly from the majority. Contrastive learning approaches, on the other hand, enforce representation similarity between positive pairs (e.g., nodes from the same neighborhood or different augmentations of the same node) while pushing apart negative pairs (e.g., nodes from different neighborhoods). Different works vary in how positive and negative pairs are defined, but the core principle remains to highlight anomalies as those nodes that fail to align with learned similarity patterns. Other approaches, such as \cite{luo2022deep}, exploit both methods.

Current state-of-the-art models adopt classic GNNs that map nodes and edges in an Euclidean latent space. Lately, scholars are exploring non-Euclidean geometries such as Hyperbolic. Such space, characterized by its constant negative curvature, diverges from the flatness of Euclidean geometry. The hyperboloid manifold is often favored for its numerical stability~\citep{fu2024hc} and for its capability in preserving hierarchical relationships~\citep{nickel2018learning}. Some approaches tried to combine different spaces to enhance their benefits~\citep{iyer2022dual, gu2018learning, cho2023curve}.

In this regard, we aim at improving the detection capabilities of NAD models by adopting a multi-geometry hybrid framework composed of a Graph-Autoencoder equipped with a contrastive learning objective. The idea is to: (i) combine spaces with diverse metric structures, i.e., Euclidean and Hyperbolic latent spaces, to highlight different complementary structural and semantic features that may be emphasized differently depending on the underlying geometry; and (ii) adopt one GAE for each geometry, eventually combining the different node representations to reconstruct the original features and to align different views of the same node, accordingly to the Graph Autoencoder and Graph Contrastive Learning approaches, respectively. To the best of our knowledge, this is the first attempt at doing this.

To summarize, our main contributions are threefold:
\begin{itemize}[leftmargin=*]
\item We propose \themodel, a novel multi-geometry Graph Autoencoder augmented with a contrastive learning objective, specifically designed for node-level graph anomaly detection.
\item We introduce, to the best of our knowledge, the first framework that jointly exploits Euclidean and Hyperbolic latent spaces for node-level graph anomaly detection, thereby capturing complementary structural and semantic patterns that are overlooked by single-geometry models.
\item We conduct an extensive experimental evaluation across four real-world datasets, demonstrating that \themodel\ consistently outperforms state-of-the-art baselines.
\end{itemize}

The paper is organized as follows. Section~\ref{sec:related_works} reviews existing work on node-level graph anomaly detection. Section~\ref{sec:geometric_preliminaries} introduces the necessary background on hyperbolic and hybrid geometric spaces. Section~\ref{sec:approach} details the proposed method. Section~\ref{sec:experiments} presents the empirical evaluation against baseline models. Finally, Section~\ref{sec:conclusion} concludes the manuscript.

% \begin{enumerate}
%     \item Motivations for hyperbolic embeddings, graphs and complex networks:\\
%     Hyperbolic space, defined by its constant negative curvature, diverges from the flatness of Euclidean geometry. The hyperboloid manifold is often favored for its numerical stability, making it a popular choice in hyperbolic geometry applications \cite{fu2024hc}.\\
%     Due to its geometric properties, hyperbolic space can be thought of as continuous analogue to \textbf{discrete trees}. By embeddings concepts in such a way that their similarity order is preserved, we can then identify (soft) hierarchical relationships from the embedding: relatedness is captured via the distance in the embedding space, while generality is captured via the norm of the embeddings \cite{nickel2018learning}.
%     % \item Anomaly detection and hyperbolic geometry, "scale invariance"
%     % \item Code, dataset and problem definition 
%     % \item Initial toy test, line graph?
% \end{enumerate}

\section{Related Works}\label{sec:related_works}
\paragraph{\textbf{Node‑Level Anomaly Detection with GNNs.}}
A rich body of work explores node-level anomaly detection on graphs using GNNs. Early methods like ANOMALOUS~\citep{anomalous} jointly learned representations and selected important attributes via CUR decomposition to mitigate noisy features. Reconstruction-based approaches, such as DOMINANT~\citep{dominant}, AnomalyDAE~\citep{fan2020anomalydae}, Radar~\citep{li2017radar}, and GAD‑NR~\citep{roy2024gad}, leverage GNN/autoencoder frameworks to detect anomalies via high reconstruction error in either structure or attributes. 

Contrastive and generative techniques refine this paradigm. GRADATE~\citep{duan2023graph} and NLGAD~\citep{duan2023normality} employ multi‑scale contrastive learning—subgraph vs. subgraph, node vs. context—to learn robust normality representations that highlight deviations. PREM~\citep{pan2023prem}, with a lightweight preprocessing-and-matching pipeline, achieves competitive detection efficiency and simplicity. SmoothGNN~\citep{dong2025smoothgnn} identifies smoothing patterns (ISP/NSP) to isolate anomalous representations explicity. Truncated Affinity Maximization~\citep{qiao2023truncated} introduces one‑class homophily modeling: it iteratively truncates non‑homophilic edges to learn tight affinity for normal nodes.  
Other works aim at improving the neighbor selection, such as Reinforcement Neighborhood Selection (RAND)~\citep{bei2023reinforcement}, using RL to improve anomaly scoring.

\paragraph{\textbf{Hyperbolic Graph Neural Networks.}}  
Hyperbolic GNNs leverage non-Euclidean geometry to effectively model hierarchical and scale-free structures in graphs. Though their application to node-level anomaly detection remains unexplored, recent works highlight their potential. For example, \cite{gu2024three} revisits anomaly detection through the lens of hyperbolic neural networks, showing improved detection. Comparative studies~\citep{touahria2024comparing} in cybersecurity benchmark Poincaré and Lorentz embedding models, demonstrating enhanced anomaly separability under hyperbolic distances. At the graph level, HC‑GLAD~\citep{fu2024hc} employs dual hyperbolic contrastive learning to distinguish anomalous graphs.

\paragraph{\textbf{Combination of different spaces.}}
Recent work has shown that embedding data in different spaces can improve representation quality. \cite{iyer2022dual} introduces a dual-geometry embedding scheme that assigns different portions of a knowledge graph to different geometric spaces, demonstrating improved modeling of two-view KGs by explicitly separating regions with distinct geometric priors. \cite{gu2018learning} proposes learning embeddings in a mixed space composed of Spherical, Euclidean, and Hyperbolic components, jointly learning both the embeddings and the curvature of each component. Similarly, in \cite{cho2023curve} the authors propose to apply such a procedure to the attention head of Graph Transformer, allowing the model to learn appropriate curvatures and exploit Euclidean, hyperbolic, and spherical components within a unified, attention-based encoder. 
The idea of using multiple curvature experts has been explored recently in HELM~\citep{he2025helm}. The authors propose training Large Language Models (LLMs) components that operate in distinct curvature spaces (a Mixture-of-Curvature Experts), showing that combining curvature-specialized experts can capture richer semantic hierarchies than single-geometry models in LLMs.

In this work, we extend this line of research to node-level anomaly detection, introducing a contrastive, autoencoder-based framework that jointly exploits Euclidean and hyperbolic representations. To the best of our knowledge, this is the first study to integrate different geometries in this task.

\section{Geometric preliminaries}\label{sec:geometric_preliminaries}
For the reader's convenience, we will briefly summarize a few theoretical concepts that will be central to our implementation of computations in the hyperbolic setting and the comparison of embeddings in different metric spaces.  

\paragraph{\textbf{Product metric spaces.}}
For relevant definitions related to metric spaces, the reader may refer to \citep{deza2009encyclopedia}. Here, we’ll collect a few useful facts for what follows. Given a metric space \( (M, d) \) and \( k > 0 \), the new function 
\begin{equation} \label{eq:boundedmetric}
    d^\prime(x,y) = k \dfrac{d(x,y)}{1 + d(x,y)}
\end{equation}
for \( x,y \in M \) is again a distance on \( M \) whose diameter is less then \( k \), namely \( 0 \le d^\prime < k \), see \citep[p.~88]{deza2009encyclopedia}.  Furthermore, if \( (M_1,d_1) \), \( (M_2,d_2) \) are two metric spaces, the new function 
\begin{equation} \label{eq:summetric}
    d^{\prime \prime} \left( (x_1,y_1), (x_2,y_2) \right) = d_1(x_1,y_1) + d_2(x_2,y_2) 
\end{equation}
for all \( (x_1,y_1), (x_2,y_2) \in M_1\times M_2 \), it is a distance 
on \( M_1 \times M_2 \), see \citep[p.~93]{deza2009encyclopedia}. Fix \( k_1, k_2 > 0 \), it follows from Equations \ref{eq:boundedmetric}, \ref{eq:summetric} that the new function 
\begin{equation} \label{eq:finalmetric}
    d^{\prime \prime \prime} \left( (x_1,y_1), (x_2,y_2) \right) = 
    k_1 \dfrac{d_1(x_1,y_1)}{1 + d_1(x_1,y_1)} + 
    k_2 \dfrac{d_2(x_2,y_2)}{1 + d_2(x_2,y_2)}
\end{equation}
is a distance on \( M_1 \times M_2 \).

\paragraph{\textbf{Hyperbolic Setting.}}  
As a model of hyperbolic geometry, we choose the Minkowski hyperboloid (of constant negative curvature $-1$) because it offers relative numerical stability and streamlines implementation, similar to prior works~\cite{fu2024hc, gu2024three}.
First of all, consider coordinates \( (x_0,\ldots, x_d)\) of \( \mathbb{R}^{d+1} \) and define the Minkowski inner product as $\langle \mathbf{x}, \mathbf{y} \rangle_{\mathcal{L}} = -x_0 y_0 + \sum_{i=1}^{d} x_i y_i$ for vectors \( \mathbf{x}, \mathbf{y} \in \mathbb{R}^{d+1} \), then  
The $d$-dimensional hyperboloid model is defined as:
\begin{equation}
\mathbb{H}^d = \left\{ \mathbf{x} \in \mathbb{R}^{d+1} \ \middle| \ \langle \mathbf{x}, \mathbf{x} \rangle_{\mathcal{L}} = -1, \ x_0 > 0 \right\}\
\end{equation}
and the geodesic distance between two points \( \mathbf{x}, \mathbf{y} \in \mathbb{H}^d \) is given by:
\begin{align}\label{eq:hyperbolic_distance}
d_{\mathcal{L}}(\mathbf{x}, \mathbf{y}) = \operatorname{arcosh} \left( -\langle \mathbf{x}, \mathbf{y} \rangle_{\mathcal{L}} \right)\ .
\end{align}
At a point \( \mathbf{x} \in \mathbb{H}^d \), we identify the tangent space \( T_{\mathbf{x}}\mathbb{H}^d \) with the linear subspace:
\begin{equation} \label{eq:hyperbolictangent}
T_{\mathbf{x}}\mathbb{H}^d = \left\{ \mathbf{v} \in \mathbb{R}^{d+1} \ \middle| \ \langle \mathbf{x}, \mathbf{v} \rangle_{\mathcal{L}} = 0 \right\}
\end{equation}
As is common in literature, the passage from \(\mathbb{H}^d\) to a tangent space is the main technical point for generalising ordinary Euclidean operations (e.g., vector-matrix multiplication) to hyperbolic settings. (See Section for more details.)

The tool that provides the mapping from a tangent space to the hyperboloid is the so-called Exponential Map.  The mapping in the opposite direction is performed by the Logarithmic Map, which is formally defined as follows:  
\begin{itemize}
    \item \emph{Exponential Map}: fix \( \mathbf{x} \in \mathbb{H}^d \) , \( \exp_{\mathbf{x}} \colon T_{\mathbf{x}}\mathbb{H}^d \rightarrow \mathbb{H}^d \) it is defined as
    \begin{align}\label{eq:exp_map}
    \exp_{\mathbf{x}}(\mathbf{v}) = \cosh(\|\mathbf{v}\|_{\mathcal{L}}) \mathbf{x} + \sinh(\|\mathbf{v}\|_{\mathcal{L}}) \frac{\mathbf{v}}{\|\mathbf{v}\|_{\mathcal{L}}}
    \end{align}
    where \( \mathbf{v} \in T_{\mathbf{x}}\mathbb{H}^d \) and \( \|\mathbf{v}\|_{\mathcal{L}} = \sqrt{ \langle \mathbf{v}, \mathbf{v} \rangle_{\mathcal{L}} } \).
    \item \emph{Logarithmic Map}: fix \( \mathbf{x} \in \mathbb{H}^d \),  \( \log_{\mathbf{x}} \colon \mathbb{H}^d \rightarrow T_{\mathbf{x}}\mathbb{H}^d  \) is defined as
    \begin{align}\label{eq:log_map}
    \log_{\mathbf{x}}(\mathbf{y}) = d_{\mathcal{L}}(\mathbf{x}, \mathbf{y}) \cdot \frac{ \mathbf{y} + \langle \mathbf{x}, \mathbf{y} \rangle_{\mathcal{L}} \mathbf{x} }{ \| \mathbf{y} + \langle \mathbf{x}, \mathbf{y} \rangle_{\mathcal{L}} \mathbf{x} \|_{\mathcal{L}} }
    \end{align}
    where  \( \mathbf{y} \in \mathbb{H}^d \).
\end{itemize}

As is rather customary, to simplify computations and streamline implementation, see for instance \cite{fu2024hc}, we fix the tangent space at \( \mathbf{o} = (1, 0, \dots, 0) \in \mathbb{H}^d \). Under this hypothesis, Equation \ref{eq:hyperbolictangent} shows that the linear subspace \( T_{\mathbf{o}}\mathbb{H}^d \) is defined by the equation \( x_0 = 0 \). Consequently, if \( \mathbf{v} \in T_{\mathbf{x}}\mathbb{H}^d \), then \( \mathbf{v} = [0, \mathbf{v}^E] \), meaning \( \mathbf{v} \) can be written as the concatenation of \( 0 \) and a vector \( \mathbf{v}^E \in \mathbb{R}^d \). Furthermore, \( \langle \mathbf{v}, \mathbf{v} \rangle_{\mathcal{L}} = \langle \mathbf{v}^E, \mathbf{v}^E \rangle \) and \( \|\mathbf{v}\|_{\mathcal{L}} = \|\mathbf{v}^E\|_2 \), where \( \langle \cdot, \cdot \rangle \) and \( \| \cdot \|_2 \) denote the usual scalar product and the associated \( L_2 \) norm in \( \mathbb{R}^d \). In light of this discussion, if \( \mathbf{v} \in T_{\mathbf{o}}\mathbb{H}^d \), we can write the exponential map as:
\begin{align}\label{eq:origin}
\exp_{\mathbf{o}}(\mathbf{v}) = \exp_{\mathbf{o}}\left( \left[0, \mathbf{v}^{E}\right] \right) = 
\left( \cosh(\|\mathbf{v}^{E}\|_2), \ \sinh(\|\mathbf{v}^{E}\|_2) \times \frac{\mathbf{v}^{E}}{\|\mathbf{v}^{E}\|_2} \right)\ .
\end{align}

\section{Approach}\label{sec:approach}
\textbf{Problem Statement.}
%\paragraph{\textbf{Problem Statement.}}
Given an undirected graph $G=(X,E)$, where $X \in \mathbb{R}^{n\times d}$ represents the node features and $E$ denotes the set of edges, we assume the existence of a subset $\hat{X} \subseteq X$ corresponding to anomalous nodes. We further assume to have the set $Y = \{y_0, \ldots, y_n \}$, where $y_i = 1$ if $x_i \in \hat{X}$, and $y_i = 0$ vice versa.
The goal is to estimate the probability distribution $P(y|G)$, where $y \in \{0,1\}$ indicates whether the node $x$ is anomalous. To this end, we aim to learn a function $f(G) \rightarrow Y$, where $f$ assigns an anomaly score to each node $x \in X$, reflecting the likelihood of being anomalous.

\textbf{\themodel\ Framework.}
To address the aforementioned problem, we introduce a multi-graph autoencoder framework, equipped with a contrastive learning objective. Specifically, we construct two distinct views for each node, denoted by $X^s$ and $X^g$, and apply a mapping function $f$ that projects nodes into a latent space. The function $f$ maps similar node views close to each other in this space. Consequently, we expect that
$
d(x_i^s, x_i^g) > d(x_j^s, x_j^g) \quad \text{if} \quad y_i = 1 \text{ (anomalous) and } y_j = 0 \text{ (normal)},$
which reflects the difficulty of $f$ in aligning representations of anomalous node views due to their inherent dissimilarity.

\textbf{Node views.} First, we define the node $i$ views as $x^s_i= x_i\in X$, hence the node starting features, and $x^g_i = [RW_i||D_i]$ as a combination of local and global structural features~\citep{DBLP:conf/wsdm/LiuD0P23}, respectively. In particular, $D_i$ represents the one-hot encoding of node degrees~\citep{DBLP:conf/kdd/QiuCDZYDWT20, DBLP:conf/iclr/XuHLJ19}, while $RW_i$~\citep{DBLP:conf/iclr/DwivediL0BB22} is computed as follows. 
\begin{align}
    RW_i &= [T_{ii}, T_{ii}^2, \ldots, T_{ii}^{d_{rw}}] \in \mathbb{R}^{d_{rw}}\\
    T &= \tilde{A}D^{-1}
\end{align}
Where $A$ is the adjacency matrix of $G$ and \( \tilde{A} = A + I \) is the adjacency matrix with added self-connections, and $D$ is the corresponding degree matrix.

\textbf{Mapping function.}
Graph Neural Networks (GNNs) have been extensively demonstrated to be effective for graph representation learning tasks. Traditional GNNs embed nodes into a Euclidean latent space, a geometric space where Euclidean distance and other classical geometric properties hold. More recently, Hyperbolic Graph Neural Networks (HGNNs)~\citep{wang2021knowledge, liu2021medical, peng2021hyperbolic, bai2023hgwavenet} have gained attention for their ability to model hierarchical and complex structures, achieving promising results across various domains.

In our approach, we leverage both Euclidean and non-Euclidean (hyperbolic) GNNs, denoted respectively as $\mathrm{GNN}^e$ and $\mathrm{GNN}^h$. The key idea is to represent nodes in multiple geometric spaces, thereby capturing complementary structural and semantic features that may be emphasized differently depending on the underlying geometry. In particular, we adopt an autoencoder-like architecture consisting of $\mathrm{GNN}^e_{enc}$ and $\mathrm{GNN}^h_{enc}$ for encoding, and $\mathrm{GNN}^e_{dec}$ and $\mathrm{GNN}^h_{dec}$ for decoding.

\textbf{Euclidean GNN.}
$\mathrm{GNN}^e_{enc}$ generates node embeddings through a message-passing architecture, which aggregates neighbors node features. The formula for generating the set of nodes embeddings $H \in \mathbb{R}^{n \times k}$ for the l-th layer is described as follows.

\begin{align}\label{eq:gnn_convolution}
    H_{(l+1)} = \sigma\left( \tilde{D}^{-1/2} \tilde{A} \tilde{D}^{-1/2} H_{(l)} W_{(l)} \right)
\end{align}
Where \( H_{(0)} = X \), \( W_{(l)} \in \mathbb{R}^{D_l \times D_{l+1}} \) is a trainable weight matrix for layer \( l \). In particular, we define as $H_{(l+1)}^{s}$ and $H_{(l+1)}^{g}$ when $H_{(0)}^{s} = X^s$ and $H_{(0)}^{s} = X^s$, respectively.

\textbf{From Euclidean GNN to Hyperbolic GNN.}
Euclidean GNNs use standard vector operations and inner products in \( \mathbb{R}^d \). Hyperbolic GNNs map data in \( \mathbb{H}^d \) and use exponential or logarithmic mappings to move between the manifold and tangent spaces. This allows them to perform hyperbolic versions of the corresponding Euclidean operations.  In particular, following Equations~\ref{eq:origin} and~\ref{eq:exp_map}, we map each $x^s \in X^s$ in the hyperbolic model $\mathbb{H}^d$:
\begin{align}
    \hat{x}^s &= \exp_{\mathbf{o}}\left(\left[0, x^s\right]\right)
\end{align}
The same logic applies for $x^g \in X^g$. We denote $\hat{X}^s = \{\hat{x}^s_0, \ldots,\hat{x}^s_n\}$ and $\hat{X}^g = \{\hat{x}^g_0, \ldots,\hat{x}^g_n\}$ as the sets of node features in the different views mapped in the hyperbolic model.

The graph convolutions in Hyperbolic GNNs, namely $\mathrm{GNN}^h_{enc}$, are applied within the tangent space. As such, $\hat{X}^s$ and $\hat{X}^g$ are first mapped into the tangent space using Equation~\ref{eq:log_map}, then we can apply the HGNN convolution:
\begin{align}\label{eq:hgnn_convolution}
    \hat{H}_{(l+1)} = \sigma\left( \tilde{D}^{-1/2} \tilde{A} \tilde{D}^{-1/2} \hat{H}_{(l)} W_{(l)} \right)
\end{align}
Similarly to Equation~\ref{eq:gnn_convolution}, we define as $\hat{H}_{(l+1)}^{s}$ and $\hat{H}_{(l+1)}^{g}$ when $\hat{H}_{(0)}^{s} = \hat{X}^s$ and $\hat{H}_{(0)}^{s} = \hat{X}^s$, respectively.

\textbf{Contrastive Learning with Product Metric.}
Following prior works, we adopt a node-level contrastive loss which maximizes the agreement between different views of the same node both in Euclidean and Hyperbolic spaces. In particular, the loss function can be formalized as follows.
\begin{align}\label{eq:cl_loss}
    \mathcal{L}_{cl} = \frac{1}{2n} \sum_{i=0}^n l_1 \left( h^g_i, \hat{h}^g_i, h^s_i, \hat{h}^s_i \right) + l_2 \left( h^g_i, \hat{h}^g_i, h^s_i, \hat{h}^s_i \right)
\end{align}

\begin{align}\label{eq:node_contrast}
    l_1 \left( h^g_i, \hat{h}^g_i, h^s_i, \hat{h}^s_i \right) = -\log\frac{e^{\left( -\mathcal{D}(h^g_i, \hat{h}^g_i, h^s_i, \hat{h}^s_i) / \tau\right)}}{\sum_{j \in \{0, \ldots, n\},\ j \ne i} e^{\left( -\mathcal{D}(h^g_i, \hat{h}^g_i, h^s_j, \hat{h}^s_j) / \tau\right)}}\ .
\end{align}

\begin{align}\label{eq:node_contrast_2}
    l_2 \left( h^s_i, \hat{h}^s_i, h^g_i, \hat{h}^g_i \right) = -\log\frac{e^{\left( -\mathcal{D}(h^s_i, \hat{h}^s_i, h^g_i, \hat{h}^g_i) / \tau\right)}}{\sum_{j \in \{0, \ldots, n\},\ j \ne i} e^{\left( -\mathcal{D}(h^s_i, \hat{h}^s_i, h^g_j, \hat{h}^g_j) / \tau\right)}}\ .
\end{align}

% \begin{align}\label{eq:node_contrast}
%     l \left( h^g_i, \hat{h}^g_i, h^s_i, \hat{h}^s_i \right) = -\log\frac{e^{\left( -\mathcal{D}(h^g_i, \hat{h}^g_i, h^s_i, \hat{h}^s_i) / \tau\right)}}{\sum_{j \in \{0, \ldots, n\},\ j \ne i} e^{\left( -\mathcal{D}(h^g_i, \hat{h}^g_i, h^g_j, \hat{h}^g_j) / \tau\right)}}\ .
% \end{align}

% \begin{align}\label{eq:node_contrast_2}
%     l \left( h^s_i, \hat{h}^s_i, h^g_i, \hat{h}^g_i \right) = -\log\frac{e^{\left( -\mathcal{D}(h^s_i, \hat{h}^s_i, h^g_i, \hat{h}^g_i) / \tau\right)}}{\sum_{j \in \{0, \ldots, n\},\ j \ne i} e^{\left( -\mathcal{D}(h^s_i, \hat{h}^s_i, h^s_j, \hat{h}^s_j) / \tau\right)}}\ .
% \end{align}

\begin{comment}
The Product Metric function is used to combine distances $g_1, \ldots, g_n$ between points $\vec{a} = ( a_1, \ldots, a_m)$ and $\vec{b} = ( b_1, \ldots, b_m)$ in different spaces~\citep{deza2009encyclopedia}:
\begin{align}
    (g_1 \times \ldots \times g_n)\left(\vec{a}, \vec{b}\right) := \sum_{i=1}^m \frac{1}{2^i}\frac{g_i(a_i, b_i)}{1+g_i(a_i, b_i)}
\end{align}
\end{comment}

In particular, in our approach, given four quantites $\{\alpha_1, \alpha_2, \alpha_3, \alpha_4\}$, we define the function $\mathcal{D}$ using Equation \ref{eq:product_metric} between views in Euclidean and Hyperbolic spaces:
% \begin{align}\label{eq:product_metric}
%     \mathcal{D}(h^g_i, \hat{h}^g_i, h^s_i, \hat{h}^s_i) = \dfrac{1}{2} \times \left( \frac{d(h^g_i, h^s_i)}{1+d(h^g_i, h^s_i)} +  \frac{\hat{d}(\hat{h}^g_i, \hat{h}^s_i)}{1+\hat{d}(\hat{h}^g_i, \hat{h}^s_i)} \right) 
% \end{align}
\begin{equation}
    \mathcal{D}(\alpha_1, \alpha_2, \alpha_3, \alpha_4) = \dfrac{1}{2} \left( \frac{d(\alpha_1, \alpha_3)}{1+d(\alpha_1, \alpha_3)} +  \frac{\hat{d}(\alpha_2, \alpha_4)}{1+\hat{d}(\alpha_2, \alpha_4)} \right)
\label{eq:product_metric}
\end{equation}
where $d$ is the \( L_2 \) norm and $\hat{d}$ is the geodesic distance in the Hyperbolic space defined in Equation~\ref{eq:hyperbolic_distance}. Observe that the right hand side in parentheses of the above formula is bounded by \( 2 \) we then divide by two to normalize \( \mathcal{D} \).

\textbf{Decoder Module.} After generating the node embeddings both in Euclidean and Hyperbolic spaces, we aim at reconstructing the input graph $G$, thus its adjacency matrix $A$ and the original and structural node features $X^s$ and $X^g$, respectively.
As prior work, we generate the reconstructed adjacency matrix $\mathcal{A}$ for original and structural features as follows.
\begin{align}
    \mathcal{A}^s &= sigmoid(H^{s}H^{s^{\top}})\\
    \mathcal{A}^g &= sigmoid(H^{g}H^{g^{\top}})\
\end{align}
Similarly, for the Hyperbolic embeddings:
\begin{align}
    \hat{\mathcal{A}^s} &= sigmoid(\hat{H}^{s}\hat{H}^{s^{\top}})\\
    \hat{\mathcal{A}^g} &= sigmoid(\hat{H}^{g}\hat{H}^{g^{\top}})\
\end{align}
Hence, we can define the reconstruction loss for the adjacency matrix:
\begin{align}
    \mathcal{L}_{adj} = ||A-\mathcal{A}^s||^2 + ||A-\mathcal{A}^g||^2 + ||A-\hat{\mathcal{A}^s}||^2 + ||A-\hat{\mathcal{A}^s}||^2
\end{align}

For reconstructing node features, we adopt two GNNs which we define as $\mathrm{GNN}^e_{dec}$ and $\mathrm{GNN}^h_{dec}$ for Euclidean and Hyperbolic embeddings:

\begin{align}\label{eq:gnn_convolution_dec}
    \mathcal{H}_{(l+1)} = \sigma\left( \tilde{D}^{-1/2} \tilde{A} \tilde{D}^{-1/2} \mathcal{H}_{(l)} W_{(l)} \right)
\end{align}

\begin{align}\label{eq:hgnn_convolution_dec}
    \hat{\mathcal{H}}_{(l+1)} = \sigma\left( \tilde{D}^{-1/2} \tilde{A} \tilde{D}^{-1/2} \hat{\mathcal{H}}_{(l)} W_{(l)} \right)
\end{align}

Here, we define the reconstructed starting and generated node features as $\mathcal{H}^s$ and $\mathcal{H}^g$, respectively. The result of Eq~\ref{eq:gnn_convolution_dec} is $\mathcal{H}^s$ whether $\mathcal{H}^s_{(0)} = H^{s}$, and $\mathcal{H}^g$ whether $\mathcal{H}^g_{(0)} = H^{g}$. The same applies for $\hat{\mathcal{H}^s}$ and $\hat{\mathcal{H}}^g$. 

For computing the node features reconstruction error in Euclidean and Hyperbolic spaces, namely $X^s, X^g, \hat{X}^s, \hat{X}^g$, we adopt, similar to Eq.~\ref{eq:product_metric}, the product metric formula to combine distances:
\begin{equation}
%\begin{aligned}
    \mathcal{L}_{node} = \dfrac{1}{2} \times \left(\frac{d(X^g, \mathcal{H}^g) + d(X^s, \mathcal{H}^s)}{1+d(X^g, \mathcal{H}^g) + d(X^s, \mathcal{H}^s)} + \frac{\hat{d}(\hat{X}^g, \hat{\mathcal{H}^g}) + \hat{d}(\hat{X}^s, \hat{\mathcal{H}^s})}{1+\hat{d}(\hat{X}^g, \hat{\mathcal{H}^g}) + \hat{d}(\hat{X}^s, \hat{\mathcal{H}^s})} \right)
%\end{aligned}
\end{equation}

Finally, combining all the components we define the loss function as follows.

\begin{equation}\label{eq:loss_function}
    \mathcal{L} = \mathcal{L}_{cl} + \lambda_1 \mathcal{L}_{adj} + \lambda_2 \mathcal{L}_{node}
\end{equation}

Eq.~\ref{eq:loss_function} defines the loss function used to train the model, which can also serve as an anomaly scoring mechanism. For this purpose, we omit the adjacency reconstruction term by setting $\lambda_1 = 0$.

% \paragraph{\textbf{Algorithm.}} 
The complete workflow presented in this section is summarized in Algorithm~\ref{alg:algorithm}.
\begin{algorithm}[ht!]
\caption{Euclidean–Hyperbolic Node Anomaly Detection}
\label{alg:algorithm}
\begin{algorithmic}[1]
\REQUIRE Graph $G=(V,E)$, adjacency $A$, node features $X$
\ENSURE Anomaly scores $S$ for all nodes

\STATE $X^s \gets X$ \COMMENT{Original node features}
\STATE Initialize $X^g \gets []$
\FOR{each node $i \in V$}
    \STATE $RW_i \gets ComputeRandomWalkFeatures(A, i)$
    \STATE $D_i \gets OneHotDegree(A, i)$
    \STATE $X^g[i] \gets concatenate(RW_i, D_i)$
\ENDFOR

\STATE $H^s, H^g \gets \mathrm{GNN}^e_{enc}(X^s, X^g, A)$ \COMMENT{Euclidean embeddings}

\STATE $\hat{X}^s \gets ExpMap(X^s)$
\STATE $\hat{X}^g \gets ExpMap(X^g)$
\STATE $\hat{H}^s, \hat{H}^g \gets \mathrm{GNN}^h_{enc}(\hat{X}^s, \hat{X}^g, A)$ \COMMENT{Hyperbolic embeddings}

\STATE $\mathcal{L}_{cl} \gets ContrastiveLoss(H^s,H^g,\hat{H}^s,\hat{H}^g)$

\STATE $\mathcal{A}^s, \mathcal{A}^g \gets AdjReconstruction(H^s, H^g)$
\STATE $\hat{\mathcal{A}}^s, \hat{\mathcal{A}}^g \gets AdjReconstruction(\hat{H}^s, \hat{H}^g)$
\STATE $\mathcal{H}^s, \mathcal{H}^g \gets \mathrm{GNN}^e_{dec}(H^s, H^g)$
\STATE $\hat{\mathcal{H}}^s, \hat{\mathcal{H}}^g \gets \mathrm{GNN}^h_{dec}(\hat{H}^s, \hat{H}^g)$

\STATE $\mathcal{L}_{adj} \gets AdjacencyLoss(A, \mathcal{A}^s, \mathcal{A}^g, \hat{\mathcal{A}}^s, \hat{\mathcal{A}}^g)$
\STATE $\mathcal{L}_{node} \gets FeatureLoss(H^s,H^g,\mathcal{H}^s,\mathcal{H}^g,\hat{H}^s,\hat{H}^g,\hat{\mathcal{H}}^s,\hat{\mathcal{H}}^g)$

\STATE $\mathcal{L} \gets \mathcal{L}_{cl} + \lambda_1 \mathcal{L}_{adj} + \lambda_2 \mathcal{L}_{node}$

\IF{AnomalyDetectionMode}
    \STATE $\lambda_1 \gets 0$ \COMMENT{Ignore adjacency reconstruction}
    \STATE $S \gets \mathcal{L}$ \COMMENT{Use loss as anomaly score}
\ENDIF

% \RETURN $S$
\end{algorithmic}
\end{algorithm}

Specifically, in our implementation, we employ uniform sampling to construct node neighborhoods, consistent with the strategy introduced in GraphSAGE~\citep{graphsage}. The neighborhood size is considered a tunable hyperparameter, and its optimal value is selected according to the characteristics of each dataset. For the underlying architecture of the employed GNNs, we adopt the Graph Isomorphism Network (GIN)~\citep{DBLP:conf/iclr/XuHLJ19}, given its expressive power in distinguishing graph structures.

\section{Experiments}\label{sec:experiments}
This section provides a comprehensive empirical evaluation of the proposed model on four diverse real-world datasets. To gain a deeper understanding of its effectiveness, we investigate its behavior through the following guiding research questions:

\begin{enumerate}[label=\textbf{RQ\arabic*}, leftmargin=*]
\item How does \themodel\ compare to state-of-the-art anomaly detection methods?
\item How do the different architectural components of \themodel\ contribute to its final performance?
\end{enumerate}

\subsection{Experimental Setup} 
Here we further detail how we devised the experiments in terms of datasets, competitors and evaluation protocols.
%To replicate and reuse the proposed approach, we publicly release our source code\footnote{GITHUB SOURCE CODE}. Our implementation is developed using the PyTorch Geometric framework~\cite{pytorchgeom1, pytorchgeom2}. All experiments were performed on an NVIDIA DGX equipped with 4 V100 GPUs.

\paragraph{\textbf{Datasets.}}
For our empirical evaluation of \themodel, we consider four real-world datasets: Disney~\cite{disney_books_dataset}, Books~\cite{disney_books_dataset}, Reddit~\cite{reddit_dataset_1, reddit_dataset_2}, and T-Finance~\cite{tfinance_dataset}.
A key characteristic of these datasets is that they contain naturally occurring (organic) anomalies, which better reflect the complexity of real-world environments. This differs from common benchmark practices where synthetic anomalies are generated by injecting artificial perturbations into clean graphs, such as Cora~\cite{cora_dataset} dataset. While convenient, such anomalies are often trivially separable from normal patterns and fail to capture the subtleties of real-world anomaly detection.
Comprehensive dataset statistics, including numbers of nodes, edges, features, and anomaly ratios, are reported in Table~\ref{tab:datasets}.

\begin{table}[!ht]
\centering
\begin{tabular}{@{}ccccc@{}}
\toprule
Dataset & \#Nodes & \#Edges & \#Features & \%Anomaly Ratio \\ \midrule
Disney & 124 & 335 & 28 & 4.8 \\
Books & 1,418 & 3,695 & 21 & 2.0 \\
Reddit & 10,984 & 168,016 & 64 & 3.3 \\
T-Finance & 39,357 & 21,222,543 & 10 & 4.6 \\ \bottomrule
\end{tabular}
\caption{Statistics on the employed datasets including number of nodes, edges, features, and percentage of anomalous nodes}
\label{tab:datasets}
\end{table}

\paragraph{\textbf{Competitors.}}
To benchmark the performance of \themodel, we compare it against six widely used anomaly detection methods, spanning from classical shallow approaches to recent deep graph-based model: \textbf{LOF}~\cite{lof}, \textbf{Isolation Forest}~\cite{isolation_forest}, \textbf{ANOMALOUS}~\cite{anomalous}, \textbf{DOMINANT}~\cite{dominant}, \textbf{CONAD}~\cite{conad}, \textbf{CARD}~\cite{card}. All baselines are implemented through the standardized PyGOD framework~\cite{pygod, bond}. Further details are provided in the Appendix~\ref{appendix:competitors}.

\paragraph{\textbf{Evaluation Protocol.}} We assess anomaly detection performance using two well-established metrics: ROC-AUC and Average Precision (AP). The former quantifies the model’s discriminative ability by considering both the true positive rate (correctly identified anomalies) and the false positive rate (normal nodes misclassified as anomalies). AP summarizes the Precision–Recall curve, reflecting the trade-off between precision and recall metrics across thresholds. A key advantage of both metrics is that they do not require fixing a decision threshold, which is crucial in anomaly detection tasks. Given the strong class imbalance typically observed in such settings, where anomalies represent only a small proportion of the nodes (Table~\ref{tab:datasets}), we use AP as the primary selection metric during hyperparameter tuning. To ensure robustness, we repeat each experiment using five random seeds. The reported results correspond to the mean performance together with the standard deviation, as summarized in the subsequent tables. We assess statistical significance of the results using the Wilcoxon signed-rank test with a confidence level of 90\%. In the tables, the best-performing models are highlighted in bold, and when two or more models are statistically indistinguishable, all tied results are marked in bold.

Consistent with previous studies~\cite{dominant,conad,card}, our evaluation is performed in a transductive setting. The complete list of hyperparameters employed for training \themodel\ is presented in the Appendix in Table~\ref{tab:hyperparameters}. Batch size and number of neighbors sampled are dataset-specific. For competitor models, we determine the best-performing configuration through empirical exploration of hyperparameters tailored to their architectures, specifically tuning hidden layer dimensions, dropout probabilities, and the trade-off coefficient balancing node feature and structural reconstruction losses.

% \begin{table}[h]
% \centering
% \begin{tabular}{@{}ll@{}}
% \toprule
% \multicolumn{2}{c}{\textbf{Training Hyperparameters}} \\
% \midrule
% Learning Rate  & $[0.0001, 0.001, 0.01]$ \\
% Layers         & $[3, 5]$ \\
% Hidden Channels       & $[8, 32]$\\
% $RW, DG$       & $[4, 8]$ \\
% Temperature $\tau$       & $[0.3, 0.6, 1.0]$ \\
% $\lambda_1$       & $[0.1, 0.01, 0.001]$ \\
% $\lambda_2$       & $1.0$ \\
% \bottomrule
% \end{tabular}
% \caption{Hyperparameters and settings used for the \themodel's training.}
% \label{tab:hyperparameters}
% \end{table}

\subsection{Results}
To address \textbf{RQ1}, Table~\ref{tab:experiments_rq1} presents the comparative performance of the proposed model against the considered baselines. The last row reports the relative percentage improvement with respect to the best-performing competitor, computed as the highest mean value for each dataset and metric. A timeout of 12 hours per seed was imposed for all experiments; under this constraint, CARD failed to complete training on the T-Finance dataset, for which we therefore report Out-Of-Time (OOT). Overall, the proposed approach consistently achieves superior results across datasets, yielding relative improvements of up to $+32.4\%$ in ROC-AUC and $+337.5\%$ in AP with respect to the best competitor.

\begin{table*}[ht!]
\resizebox{\columnwidth}{!}{
\begin{tabular}{@{}lcccccccc@{}}
\toprule
\multicolumn{1}{c}{\multirow{2}{*}{Model}} & \multicolumn{2}{c}{Disney} & \multicolumn{2}{c}{Books} & \multicolumn{2}{c}{Reddit} & \multicolumn{2}{c}{T-Finance} \\ \cmidrule(l){2-9} 
\multicolumn{1}{c}{} & ROC-AUC & \multicolumn{1}{c|}{AP} & ROC-AUC & \multicolumn{1}{c|}{AP} & ROC-AUC & \multicolumn{1}{c|}{AP} & ROC-AUC & AP \\ \midrule
LOF & $0.479 \pm 0.000$ & $0.052 \pm 0.000$ & $0.365 \pm 0.000$ & $0.015 \pm 0.000$ & $0.572 \pm 0.000$ & \bm{$0.042 \pm 0.000$} & $0.493 \pm 0.000$ & $0.052 \pm 0.000$ \\
Isolation Forest & $0.573 \pm 0.026$ & $0.094 \pm 0.029$ & $0.417 \pm 0.011$ & $0.018 \pm 0.002$ & $0.461 \pm 0.012$ & $0.028 \pm 0.001$ & $0.636 \pm 0.019$ & $0.056 \pm 0.003$ \\ \midrule  
ANOMALOUS & $0.518 \pm 0.000$ & $0.072 \pm 0.000$ & $0.437 \pm 0.000$ & $0.018 \pm 0.000$ & $0.469 \pm 0.022$ & $0.030 \pm 0.002$ & $0.282 \pm 0.000$ & $0.030 \pm 0.000$ \\
DOMINANT & $0.452 \pm 0.020$ & $0.080 \pm 0.026$ & \bm{$0.596 \pm 0.037$} & \bm{$0.032 \pm 0.004$} & $0.559 \pm 0.002$ & $0.038 \pm 0.000$ & $0.474 \pm 0.142$ & $0.043 \pm 0.011$ \\
CONAD & $0.495 \pm 0.031$ & $0.059 \pm 0.006$ & $0.461 \pm 0.005$ & $0.017 \pm 0.000$ & $0.555 \pm 0.003$ & $0.038 \pm 0.000$ & $0.345 \pm 0.008$ & $0.032 \pm 0.000$ \\
CARD & $0.512 \pm 0.015$ & $0.059 \pm 0.002$ & $0.470 \pm 0.018$ & $0.022 \pm 0.002$ & $0.533 \pm 0.016$ & $0.035 \pm 0.001$ & OOT & OOT \\ \midrule
$\text{\themodel}$ & \bm{$0.705 \pm 0.045$} & \bm{$0.211 \pm 0.119$} & \bm{$0.567 \pm 0.024$} & \bm{$0.038 \pm 0.010$} & \bm{$0.609 \pm 0.031$} & \bm{$0.043 \pm 0.003$} & \bm{$0.829 \pm 0.051$} & \bm{$0.284 \pm 0.117$} \\ \midrule
\% Improv. & $+23.0\%$ & $+124.5\%$ & $-4.8\%$ & $+18.8\%$ & $+6.5\%$ & $+2.3\%$ & $+32.4\%$ & $+337.5\%$ \\ \bottomrule
\end{tabular}
}
\caption{Performance comparison on anomaly detection benchmarks using ROC-AUC and AP. The last row (\% Improv.) shows the relative improvement of \themodel\ over the best-performing competitor for each metric.}
\label{tab:experiments_rq1}
\end{table*}

\begin{figure}[ht!]
    \centering

    \begin{subfigure}{0.24\textwidth}
        \centering
        \includegraphics[width=\linewidth]{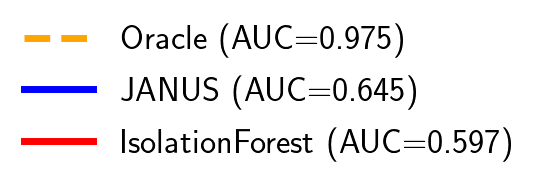}
    \end{subfigure}
    \begin{subfigure}{0.24\textwidth}
        \centering
        \includegraphics[width=\linewidth]{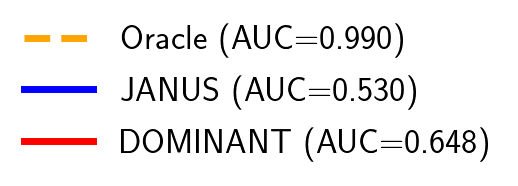}
    \end{subfigure}
    \begin{subfigure}{0.2\textwidth}
        \centering
        \includegraphics[width=\linewidth]{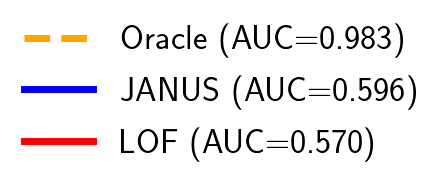}
    \end{subfigure}
    \begin{subfigure}{0.24\textwidth}
        \centering
        \includegraphics[width=\linewidth]{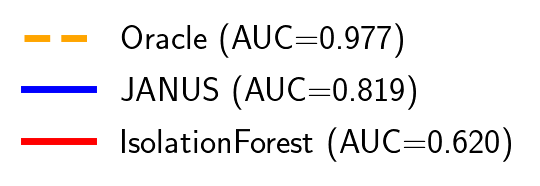}
    \end{subfigure}
    
    \begin{subfigure}[b]{.23\textwidth}
        \centering
        \includegraphics[width=\textwidth]{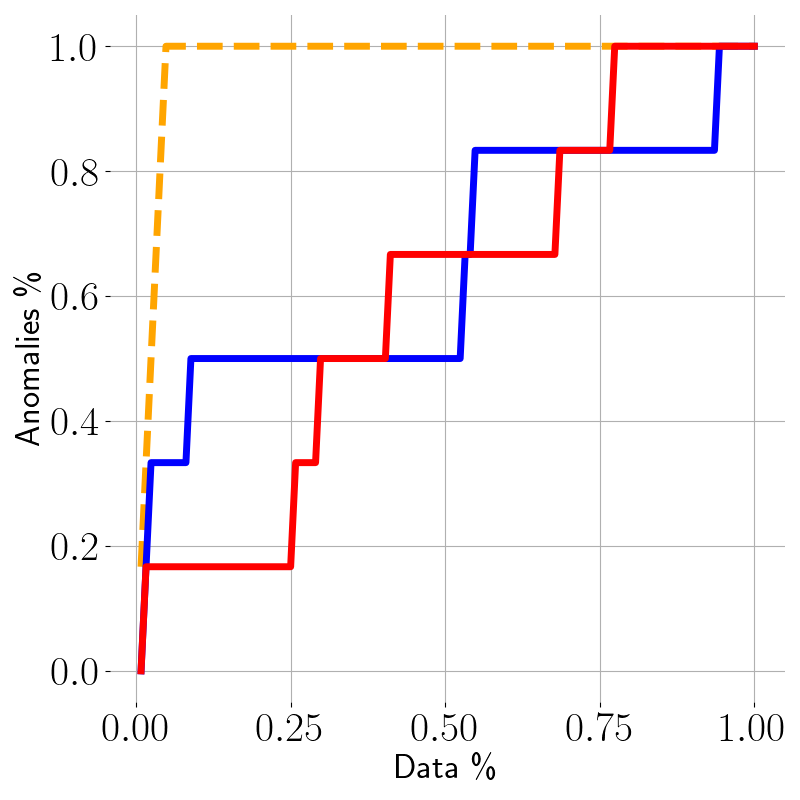}
        \caption{Disney}
    \end{subfigure}
    \begin{subfigure}[b]{.23\textwidth}
        \centering
        \includegraphics[width=\textwidth]{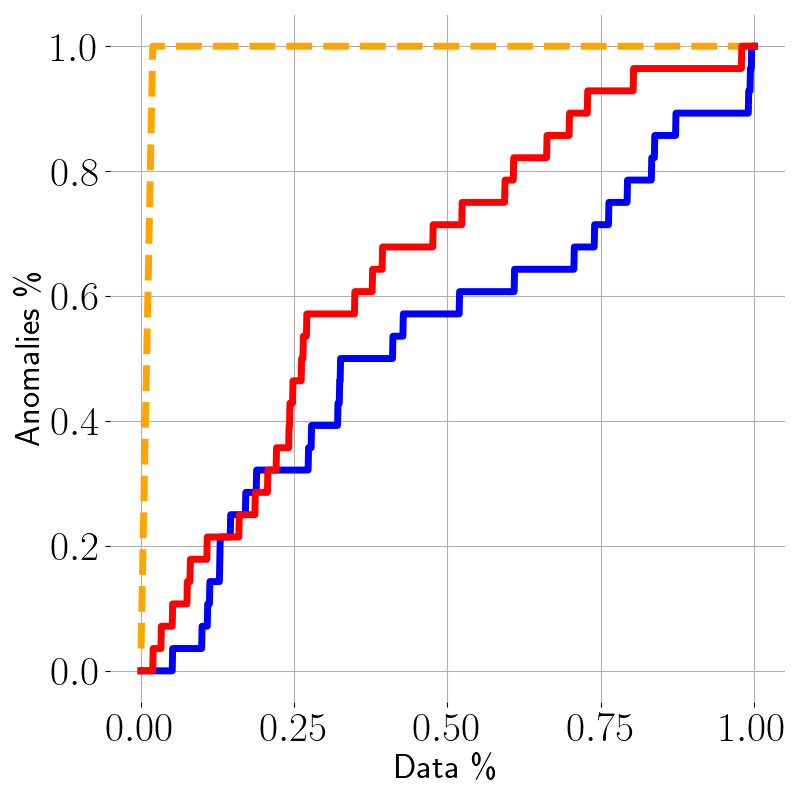}
        \caption{Books}
    \end{subfigure}
    \begin{subfigure}[b]{.23\textwidth}
        \centering
        \includegraphics[width=\textwidth]{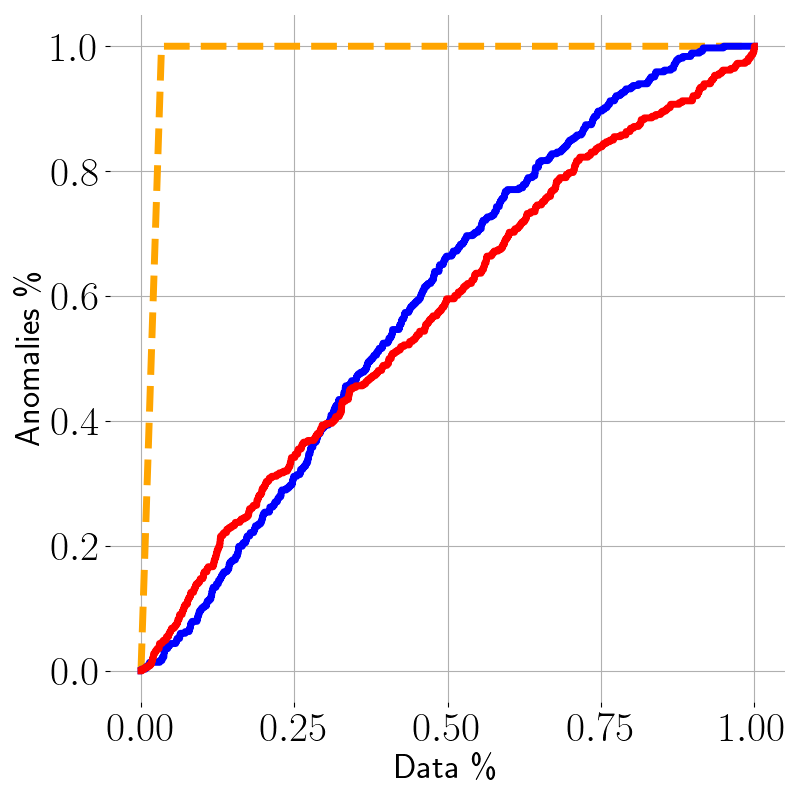}
        \caption{Reddit}
    \end{subfigure}
    \begin{subfigure}[b]{.23\textwidth}
        \centering
        \includegraphics[width=\textwidth]{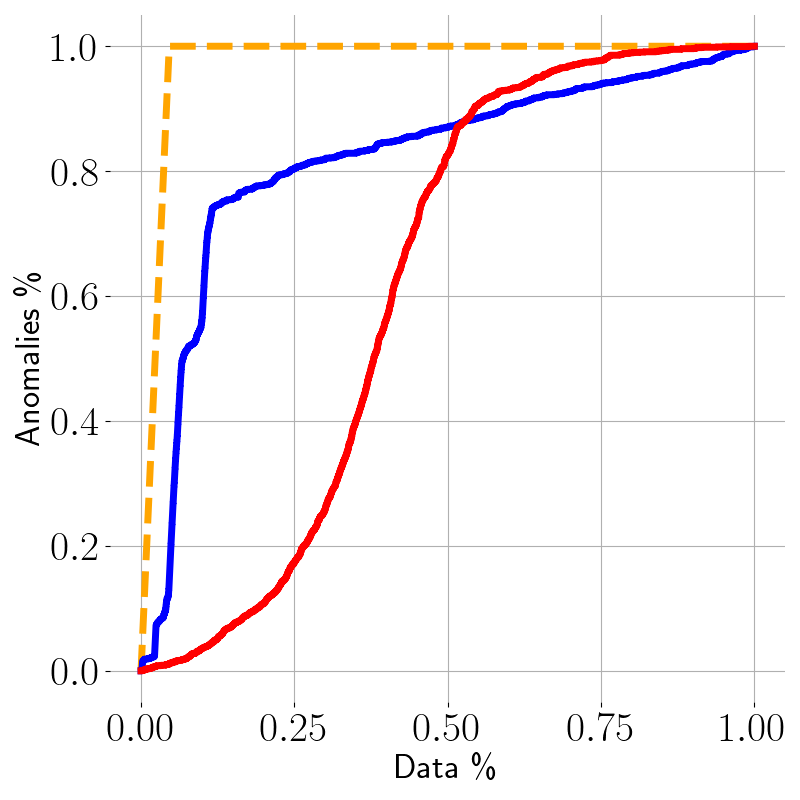}
        \caption{T-Finance}
    \end{subfigure}
    \caption{Cumulative Gain charts across datasets. The y-axis shows the percentage of detected anomalies, while the x-axis shows the increasing fraction of samples of the dataset adopted.}
    \label{fig:cumulative_gain_charts}
\end{figure}

To further strengthen the evaluation of the proposed approach, we report in Figure~\ref{fig:cumulative_gain_charts} the Cumulative Gain curves across the considered datasets. These plots illustrate the percentage of correctly detected anomalies as a function of the number of nodes analyzed, where nodes are ranked in descending order according to the anomaly scores assigned by the model. Analogously to the ROC-AUC, we also compute the Area Under the Cumulative Gain Curve (higher is better) to facilitate a direct comparison between \themodel\ and the strongest competitor identified in Table~\ref{tab:experiments_rq1}, alongside a perfect classifier (the Oracle). Results empirically show that \themodel\ consistently dominates the competitors, except for Books.

Finally, an analysis comparing the time cost of \themodel\ against its predictive quality, in relation to the selected competitors, is presented in the Appendix~\ref{app:time}.

\subsection{Ablation Study}
% \textcolor{red}{TODO, probabilmente va messa in appendice}

To address \textbf{RQ2}, Table~\ref{tab:experiments_rq2} reports the results of an ablation study on the individual components of \themodel. Specifically, we compare the complete model against two simplified variants: one leveraging only the autoencoder component, denoted as $\text{\themodel}^{AE(s,\ g)}$, and the other relying solely on the contrastive learning component, denoted as $\text{\themodel}^{CL(s,\ g)}$. The results highlight that \themodel\ achieves better stability and a more balanced performance across both ROC-AUC and AP metrics.

\begin{table*}[ht!]
\resizebox{\columnwidth}{!}{
\begin{tabular}{@{}lcccccccc@{}}
\toprule
\multicolumn{1}{c}{\multirow{2}{*}{Model}} & \multicolumn{2}{c}{Disney} & \multicolumn{2}{c}{Books} & \multicolumn{2}{c}{Reddit} & \multicolumn{2}{c}{T-Finance} \\ \cmidrule(l){2-9} 
\multicolumn{1}{c}{} & ROC-AUC & \multicolumn{1}{c|}{AP} & ROC-AUC & \multicolumn{1}{c|}{AP} & ROC-AUC & \multicolumn{1}{c|}{AP} & ROC-AUC & AP \\ \midrule
$\text{\themodel}^{AE(s,\ g)}$ & $0.685 \pm 0.045$ & $0.111 \pm 0.005$ & $0.644 \pm 0.004$ & $0.027 \pm 0.001$ & $0.574 \pm 0.017$ & $0.041 \pm 0.002$ & $0.789 \pm 0.055$ & $0.164 \pm 0.055$\\
$\text{\themodel}^{CL(s,\ g)}$ & $0.713 \pm 0.091$ & $0.183 \pm 0.054$ & $0.521 \pm 0.036$ & $0.051 \pm 0.008$ & $0.556 \pm 0.011$ & $0.039 \pm 0.001$ & $0.818 \pm 0.100$ & $0.290 \pm 0.113$\\ \midrule
% $\text{\themodel}^{AE+CL(s)}$ & $0.646 \pm 0.053$ & $0.089 + 0.011$ & $0.521 \pm 0.034$ & $0.038 \pm 0.005$ & $0.582 \pm 0.009$ & $0.047 \pm 0.001$ & $0.851 \pm 0.018$ & $0.353 \pm 0.109$\\
% $\text{\themodel}^{AE+CL(g)}$ & $0.656 \pm 0.047$ & $0.296 \pm 0.049$ & $0.560 \pm 0.020$ & $0.045 \pm 0.011$ & $0.487 \pm 0.017$ & $0.033 \pm 0.001$ & $0.793 \pm 0.098$ & $0.179 \pm 0.069$ \\ \midrule
$\text{\themodel}$ & $0.705 \pm 0.045$ & $0.211 \pm 0.119$ & $0.567 \pm 0.024$ & $0.038 \pm 0.01$ & $0.609 \pm 0.031$ & $0.043 \pm 0.003$ & $0.829 \pm 0.051$ & $0.284 \pm 0.117$ \\ \bottomrule
\end{tabular}
}
\caption{Ablation results of \themodel\ under different settings: autoencoder ($AE$), contrastive learning ($CL$), and their combinations, with starting ($s$) and generated ($g$) node features. The complete \themodel\ achieves consistently strong performance by combining Euclidean and Hyperbolic representations.}
\label{tab:experiments_rq2}
\end{table*}

%\todo[inline]{From Simone: Togliamo AE+CL(s) e AE+CL(g)?}

\section{Conclusion}\label{sec:conclusion}
In this work, we introduced Janus, a novel multi-geometry framework that integrates Euclidean and Hyperbolic representations for node-level anomaly detection. By jointly leveraging the complementary properties of both spaces within a graph autoencoder equipped with a contrastive objective, Janus captures subtle structural and semantic irregularities that single-geometry models tend to overlook. Our extensive evaluation across four real-world datasets demonstrates that Janus consistently outperforms state-of-the-art baselines, yielding significant improvements in both ROC-AUC and Average Precision. These results highlight the potential of hybrid geometric approaches to advance anomaly detection in graphs.

%Future directions include extending Janus to dynamic or heterogeneous graphs, exploring adaptive curvature learning, and investigating scalability to even larger networks.

\section*{Reproducibility Statement}
To ensure reproducibility of our results, we provide full experimental support, including the implementation of our method along with preprocessing and training scripts, available at \url{https://anonymous.4open.science/r/JANUS-5EDF/}. These resources allow other researchers to replicate our results and explore additional applications of our approach.

\section*{Acknowledgements}
This work has been partially funded by (i) Project SERICS (PE00000014) under the MUR National Recovery and Resilience Plan funded by the European Union -- NextGenerationEU; (ii) MUR on D.M. 352/2022, PNRR Ricerca, CUP H23C22000550005; (iii) Project FAIR (Future AI Research), under program NRRP MUR funded by EU-NGEU (PE00000013)
\bibliography{ref}
\bibliographystyle{plainnat}

\clearpage

\appendix

\section{Time Analysis}
\label{app:time}
\begin{figure*}[ht!]
    \centering
    
    % Legend on top
    \includegraphics[width=0.7\textwidth]{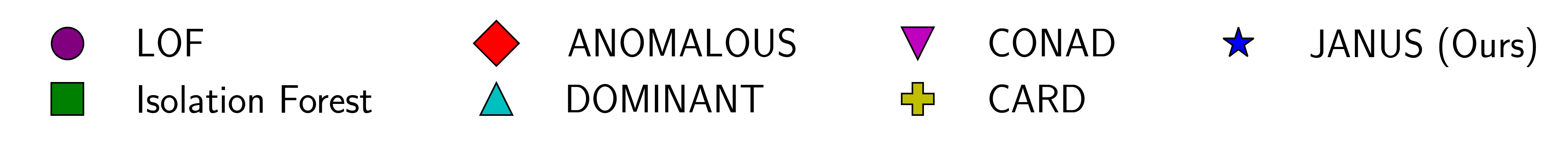}\\[1ex]
    
    % === Row 1: AP ===
    \begin{subfigure}{0.3\textwidth}
        \centering
        \includegraphics[width=\linewidth]{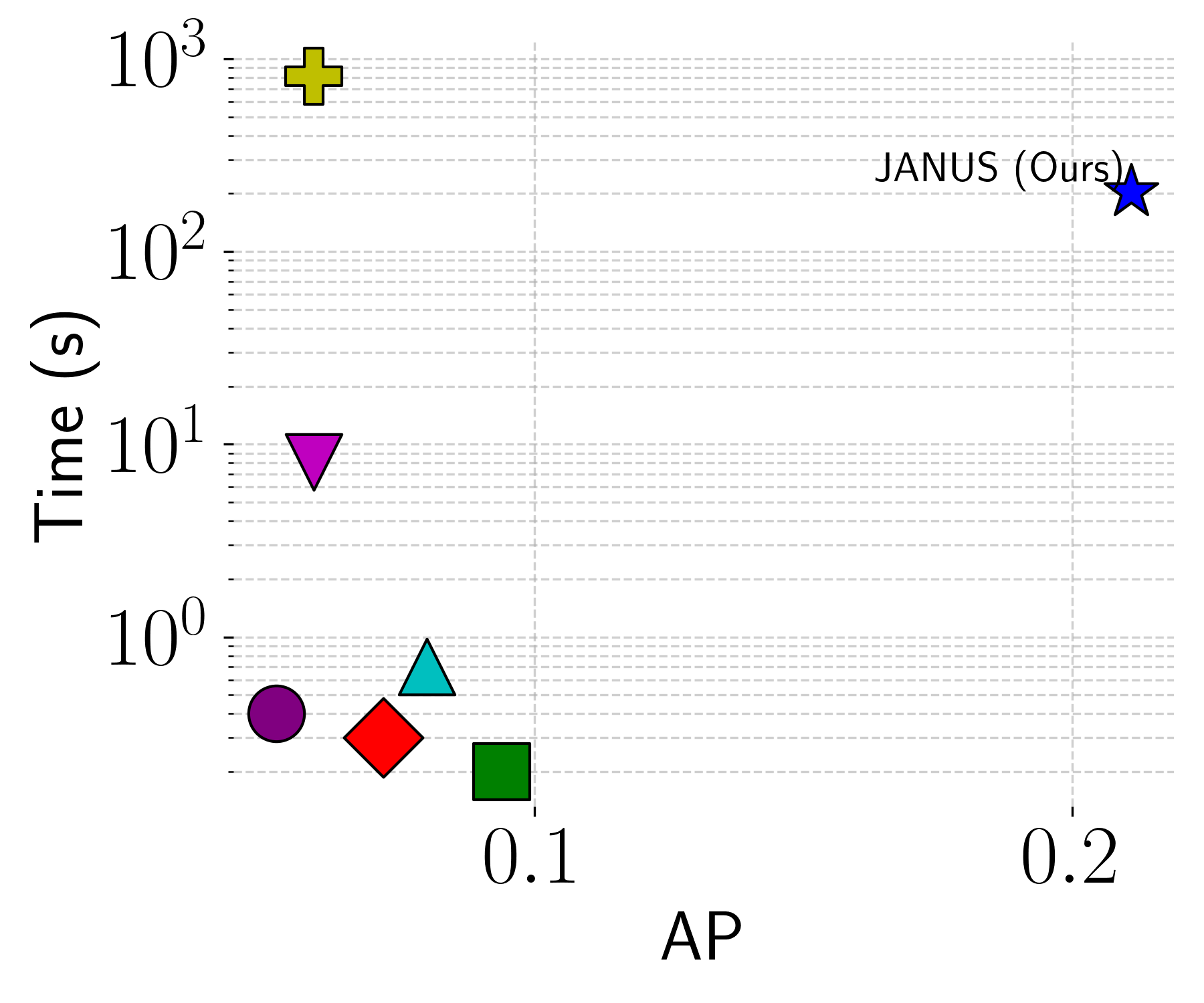}
        \caption{Disney (AP)}
    \end{subfigure}
    \qquad \qquad
    \begin{subfigure}{0.3\textwidth}
        \centering
        \includegraphics[width=\linewidth]{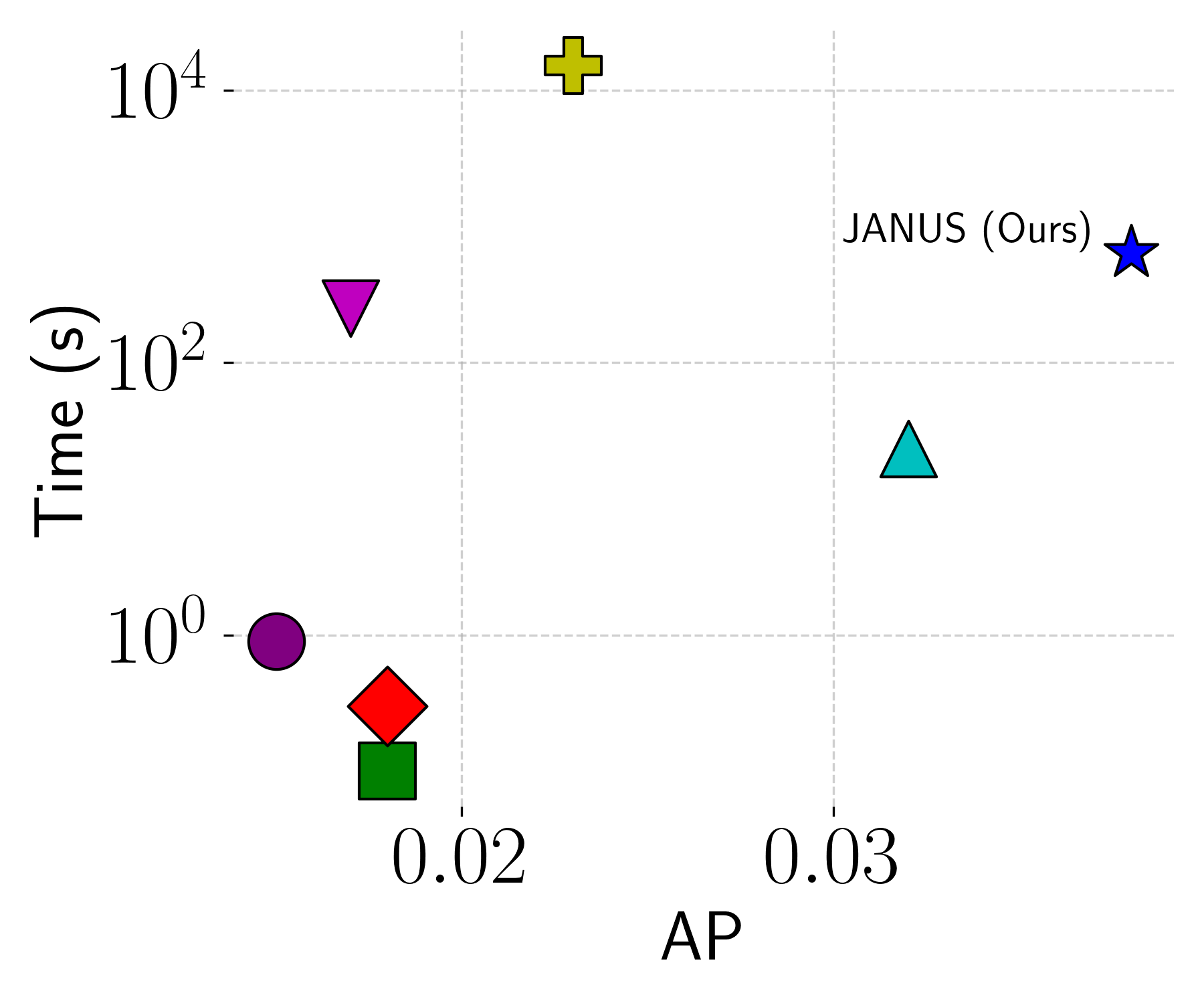}
        \caption{Books (AP)}
    \end{subfigure}
    \\
    \begin{subfigure}{0.3\textwidth}
        \centering
        \includegraphics[width=\linewidth]{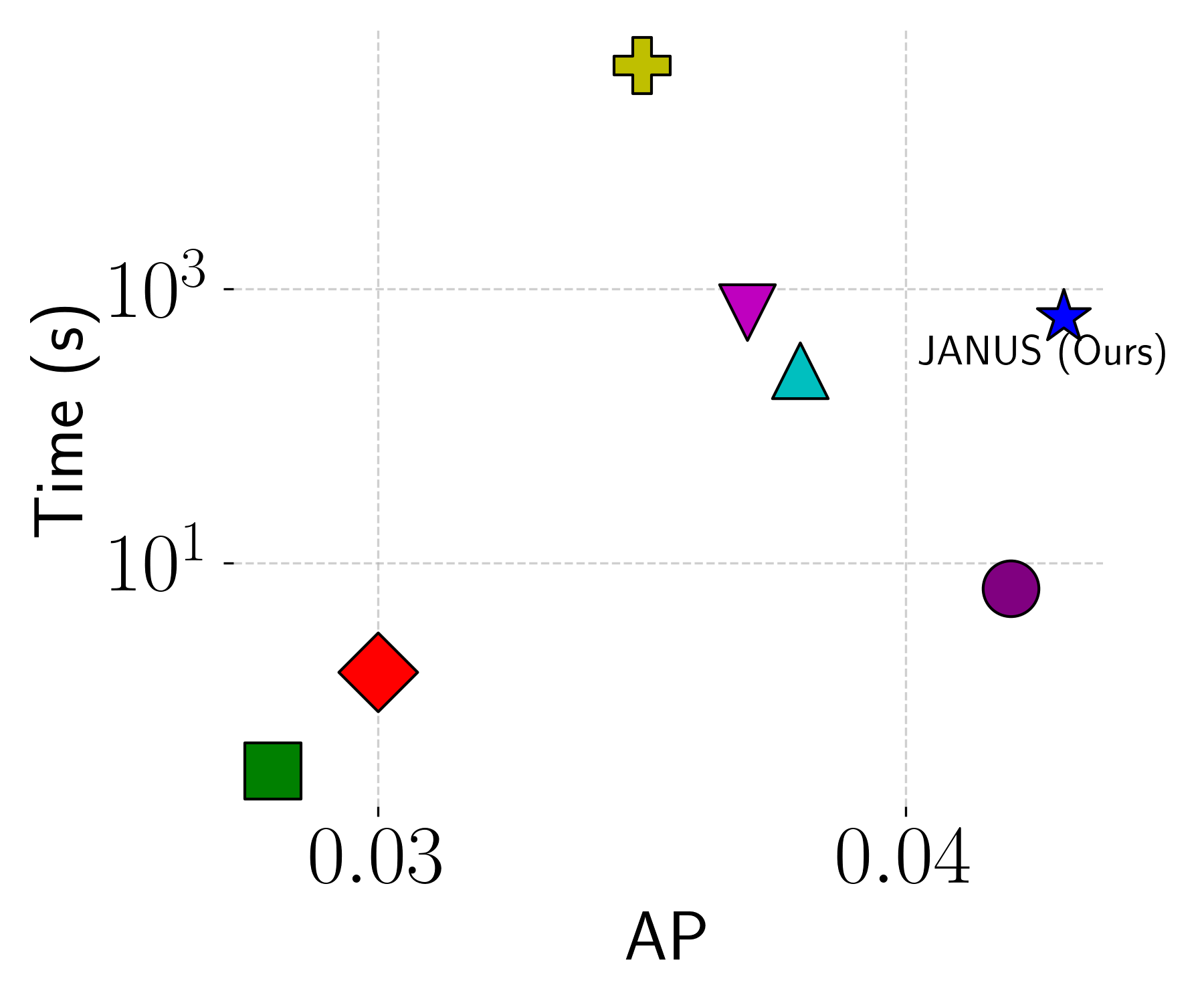}
        \caption{Reddit (AP)}
    \end{subfigure}
    \qquad \qquad
    \begin{subfigure}{0.3\textwidth}
        \centering
        \includegraphics[width=\linewidth]{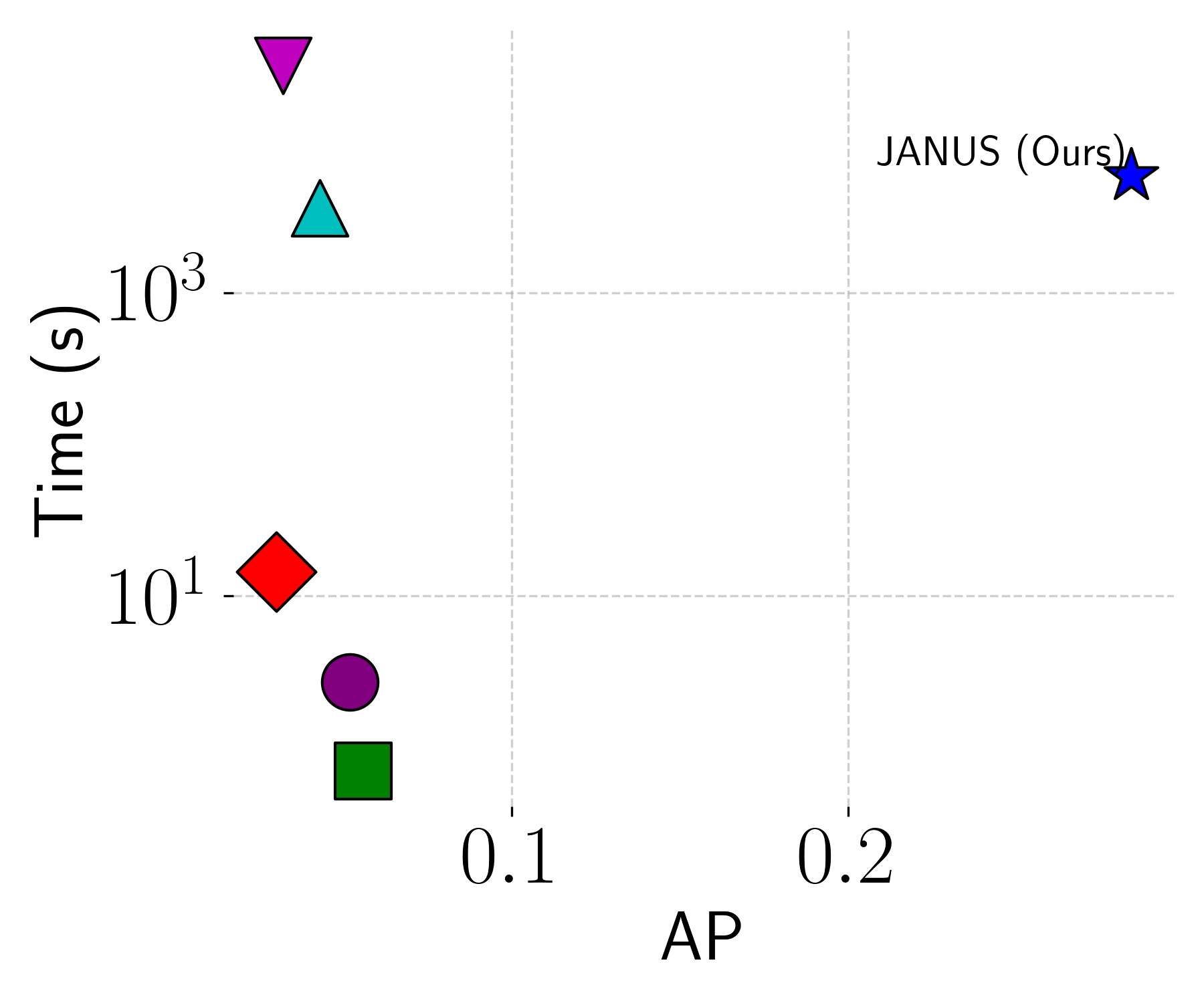}
        \caption{T-Finance (AP)}
    \end{subfigure}
    
    % === Row 2: AUC ===
    \begin{subfigure}{0.3\textwidth}
        \centering
        \includegraphics[width=\linewidth]{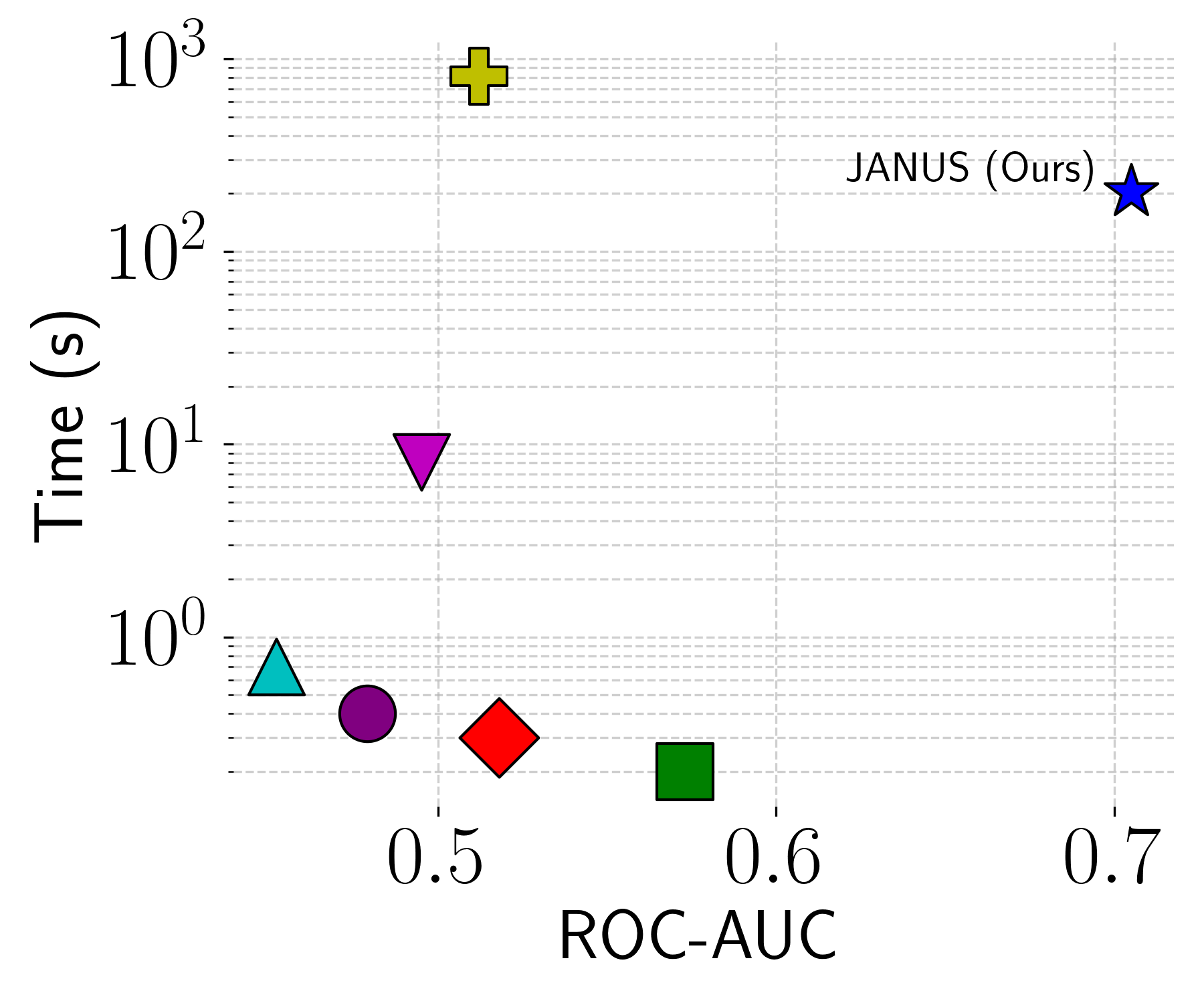}
        \caption{Disney (ROC-AUC)}
    \end{subfigure}
    \qquad \qquad
    \begin{subfigure}{0.3\textwidth}
        \centering
        \includegraphics[width=\linewidth]{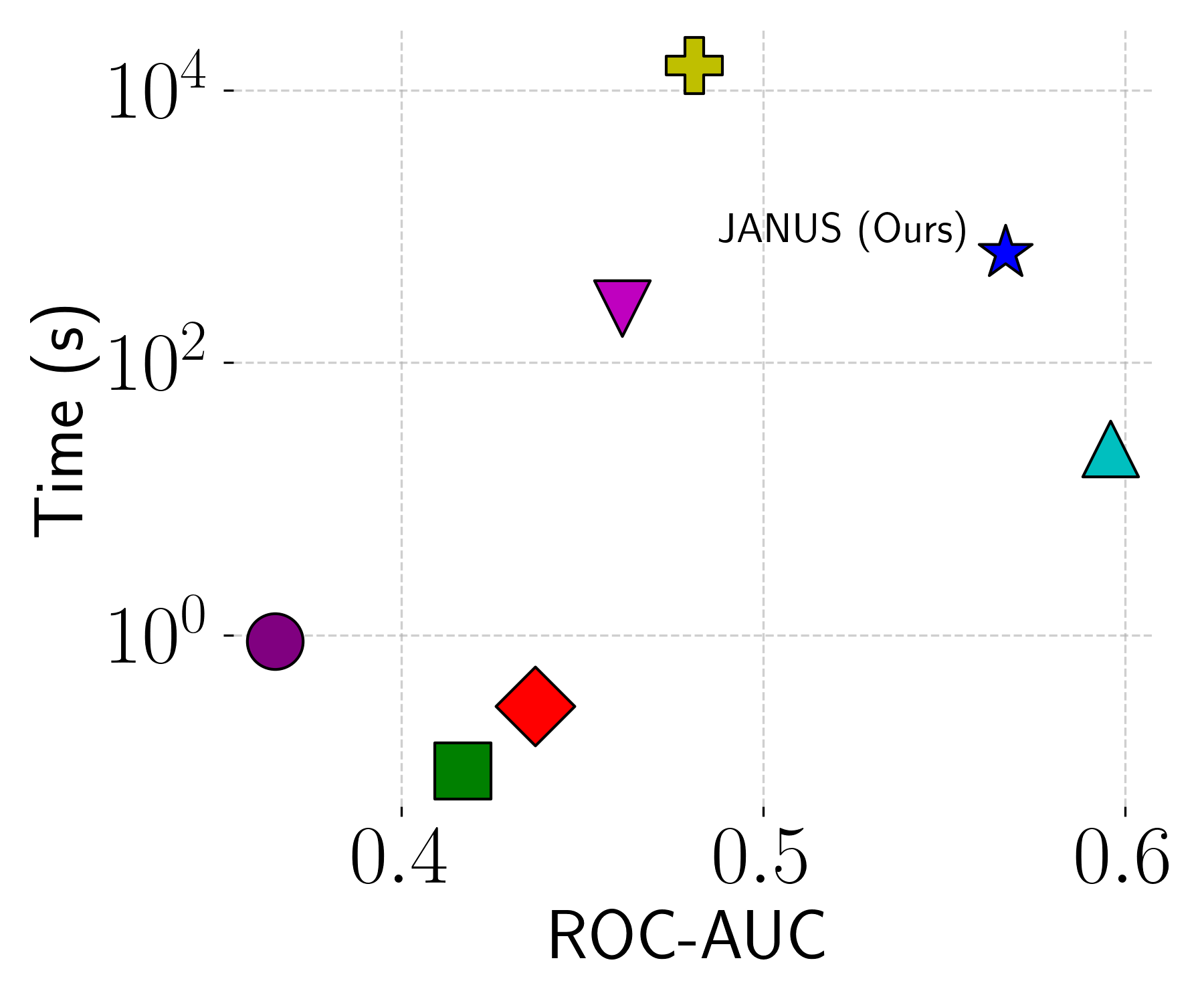}
        \caption{Books (ROC-AUC)}
    \end{subfigure}
    \\
    \begin{subfigure}{0.3\textwidth}
        \centering
        \includegraphics[width=\linewidth]{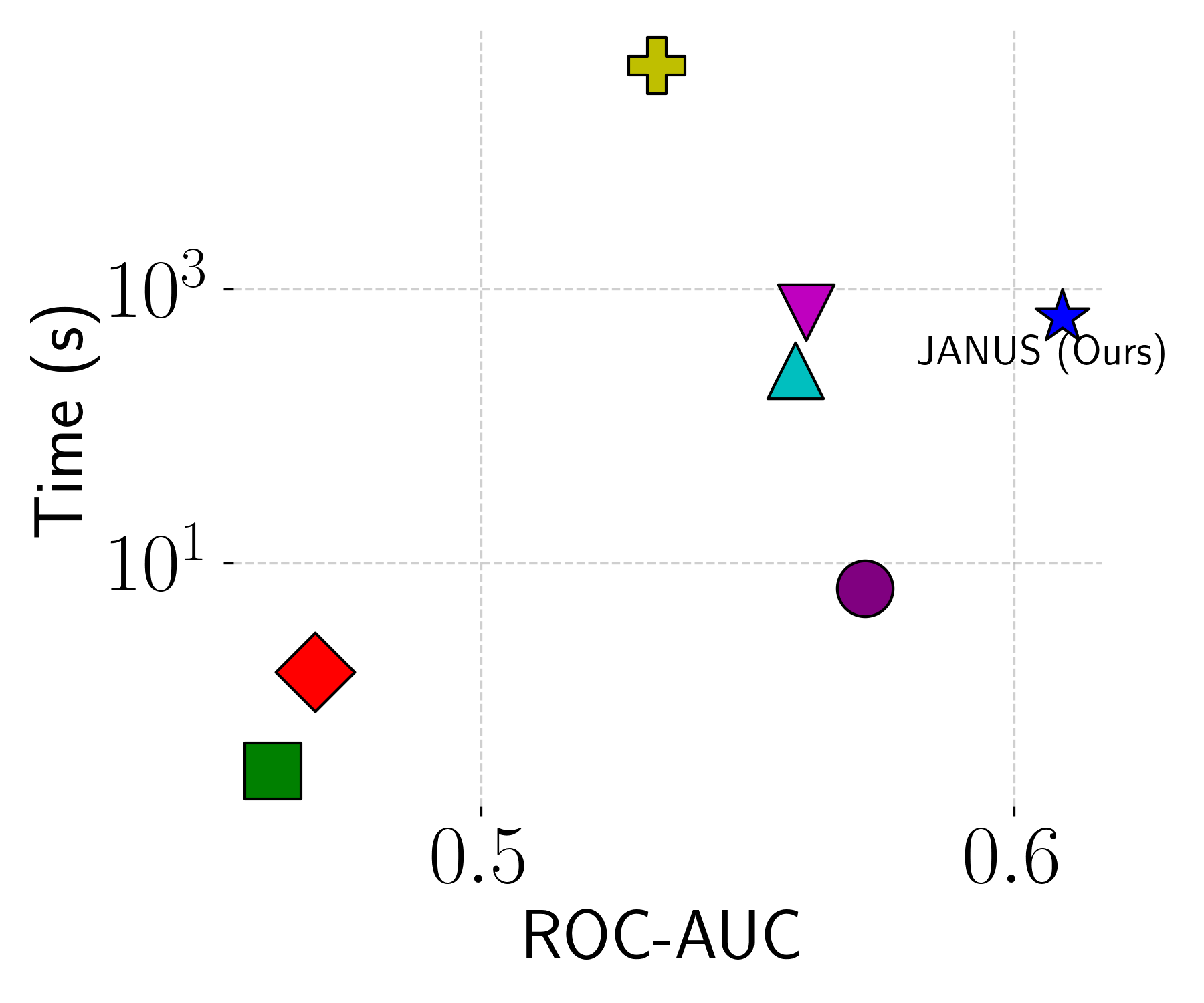}
        \caption{Reddit (ROC-AUC)}
    \end{subfigure}
    \qquad \qquad
    \begin{subfigure}{0.3\textwidth}
        \centering
        \includegraphics[width=\linewidth]{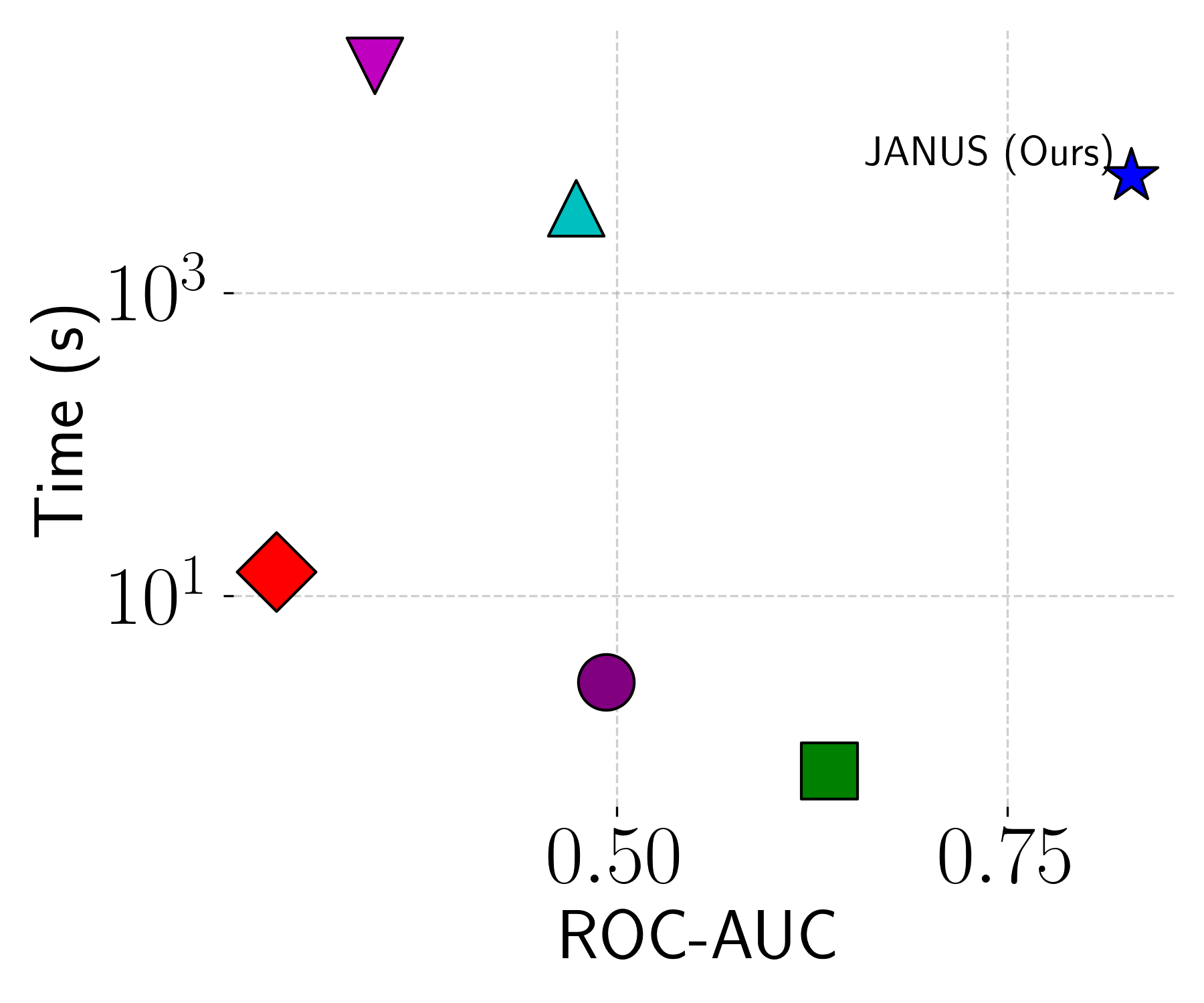}
        \caption{T-Finance (ROC-AUC)}
    \end{subfigure}

    \caption{Comparison of models across datasets. Top row: AP vs Time. Bottom row: ROC-AUC vs Time. The legend is shared across all subplots.}
    \label{fig:time_ap_auc_comparison}
\end{figure*}

Figure~\ref{fig:time_ap_auc_comparison} depicts the relationship between detection accuracy, quantified through AP and ROC-AUC, and computational cost, expressed in terms of processing time (seconds), for all the considered datasets. While some baselines either achieve lower accuracy or need high execution timings, \themodel\ demonstrates a balanced trade-off by combining competitive efficiency with consistently superior predictive performance. This highlights its practical suitability for real-world scenarios where both accuracy and scalability are crucial.

\section{Experimental Setup}

Table~\ref{tab:hyperparameters} provides the complete list of hyperparameters employed for training \themodel.

\begin{table}[h]
\centering
\begin{tabular}{@{}l l l l@{}}
\toprule
\multicolumn{4}{c}{\textbf{Training Hyperparameters}} \\
\midrule
Learning Rate       & $[0.0001, 0.001, 0.01]$ & Layers          & $[3, 5]$ \\
Hidden Channels     & $[8, 32]$               & $RW, DG$        & $[4, 8]$ \\
Temperature $\tau$  & $[0.3, 0.6, 1.0]$       & $\lambda_1$     & $[0.1, 0.01, 0.001]$ \\
$\lambda_2$         & $1.0$                    &                 &          \\
\bottomrule
\end{tabular}
\caption{Hyperparameters and settings used for the \themodel's training.}
\label{tab:hyperparameters}
\end{table}

\subsubsection*{Competitor models}\label{appendix:competitors}
\begin{itemize}
    \item \textbf{LOF}~\cite{lof}. Local Outlier Factor is a density-based anomaly detection method that identifies anomalies by comparing the local density of a node with those of its neighbors. Nodes residing in significantly lower-density regions are considered anomalous.
    \item \textbf{Isolation Forest}~\cite{isolation_forest}. This ensemble-based method isolates anomalies by recursively partitioning the data space with random decision trees. Since anomalies are more easily separated than normal points, they require fewer splits to be isolated, making this method computationally efficient and scalable.  
    \item \textbf{ANOMALOUS}~\cite{anomalous}. A pioneering shallow graph-based method, ANOMALOUS applies a CUR matrix decomposition on the attribute–structure interaction matrix of the graph. The model detects abnormal nodes by capturing inconsistencies between topology and node features.
    \item \textbf{DOMINANT}~\cite{dominant}. A deep learning–based model that extends autoencoder architectures to graphs. DOMINANT jointly reconstructs both the adjacency matrix and node attributes via GCNs, with large reconstruction errors signaling potential anomalies.
    \item \textbf{CONAD}~\cite{conad}. A contrastive learning framework designed for anomaly detection in attributed graphs. CONAD generates augmented graph views and leverages contrastive objectives to distinguish normal nodes (consistent across views) from anomalous ones (inconsistent across views).
    \item \textbf{CARD}~\cite{card}. The model refines contrastive representation learning by incorporating neighborhood-aware augmentations and anomaly-oriented regularization. Its design enables it to capture subtle irregularities in both topology and node features.  
\end{itemize}

\end{document}